\documentclass[11pt]{article}

\usepackage[margin=1in]{geometry}
\usepackage{amsmath, amssymb}
\usepackage{graphicx}
\usepackage{hyperref}
\usepackage{authblk}
\usepackage{booktabs, multirow, tabularx, caption}
\usepackage{adjustbox}
\usepackage[ruled,vlined]{algorithm2e}
\usepackage{cleveref}

\title{SAMSA 2.0: Prompting Segment Anything with Spectral Angles for Hyperspectral Interactive Medical Image Segmentation}

\author[1]{Alfie Roddan}
\author[1]{Tobias Czempiel}
\author[1]{Chi Xu}
\author[1]{Daniel S. Elson}
\author[1]{Stamatia Giannarou}

\affil[1]{\small The Hamlyn Centre for Robotic Surgery, Department of Surgery and Cancer \\
Imperial College London\\
\texttt{agr21@ic.ac.uk}}

\begin{document}

\maketitle

\begin{abstract}
We present \textbf{SAMSA~2.0}, an interactive segmentation framework for hyperspectral medical imaging that extends the Segment Anything Model~(SAM2) through \emph{spectral angle prompting}. By fusing spectral similarity cues with spatial prompts at inference time, SAMSA~2.0 injects spectral awareness into large foundation models without retraining. We evaluate SAMSA~2.0 across four hyperspectral medical datasets (HEIPOR, HIB, SB-X, and SB-H) and multiple SAM2 backbones. Compared to spectral-only baselines, SAMSA~2.0 effectively overcomes the lack of spatial modeling, while consistently outperforming RGB-only and prior fusion approaches. On average, SAMSA~2.0 achieves up to \textbf{+3.8\%} improvement in DICE@0.5 over the best RGB-only SAM2 variant and \textbf{+3.1\%} over the original SAMSA. In low-interaction settings, SAMSA~2.0 demonstrates strong robustness, with the SAMSA~2-Small variant reaching \textbf{84.2\%} DICE using a single click and up to \textbf{96.4\%} DICE with five clicks on HIB. These results highlight the effectiveness of early spectral–spatial fusion for few-shot and zero-shot hyperspectral medical image segmentation, particularly in data-limited and clinically noisy scenarios.
\end{abstract}

\bigskip
\noindent\textbf{Keywords:} Hyperspectral Imaging, Interactive Segmentation, Spectral Angle Prompting, Generalization

\section{Introduction} \label{sec:introduction}

Hyperspectral imaging (HSI) captures detailed spectral information across hundreds of bands, enabling precise tissue characterization and material identification beyond the capabilities of standard RGB imaging \cite{Clancy2020}. In clinical settings, HSI has shown significant promise in supporting critical tasks such as tumor detection and perfusion monitoring. However, the practical application of HSI remains challenging due to high data dimensionality, the scarcity of expert-annotated datasets, and significant variability across different imaging hardware \cite{Shapey2019, ANICHINI2024108293}.

Interactive segmentation has emerged as a vital tool to help clinicians extract regions of interest with minimal manual input \cite{zhao2013overview}. Traditionally, these approaches compare the spectrum of each pixel to a reference spectrum obtained from a user click, using metrics like spectral angle mapper (SAM) \cite{boardman1993spectral}. While these methods are effective at capturing intrinsic spectral features, they are often purely pixel-wise; they ignore vital spatial structures and frequently fail in noisy or complex surgical scenes.

Conversely, foundational segmentation models trained on massive RGB datasets, such as the Segment Anything Model (SAM) \cite{kirillov2023segment} and its successor SAM 2 \cite{ravi2024sam2segmentimages}, excel at spatial reasoning and zero-shot generalization. However, these models are not inherently designed to process spectral data. To bridge this gap, our previous work, SAMSA \cite{SAMSA}, combined spectral similarity maps with spatial models to reduce the need for large annotated datasets and mitigate scanner variability.

Despite its success, the original SAMSA framework had limitations: it was memory-intensive, limiting its use with larger models, and it introduced spectral information too late in the network. This late-stage integration constrained the influence of rich spectral cues on the final segmentation boundaries. Furthermore, its reliance on manual clicks highlighted a need for more automated workflows.

In this work, we introduce \textbf{SAMSA 2.0}, a refined framework designed to address these architectural and operational weaknesses. The core innovation is a method we call \emph{spectral angle prompting}---a pixel-wise similarity map derived from user clicks that is embedded directly into the model's input. By providing this information at the start, the prompt acts as a soft constraint, guiding the model with both spatial and spectral context simultaneously. This version also adapts the automatic segmentation capabilities of SAM 2 for HSI, creating a hybrid workflow where the system generates expert-verifiable proposals.

The main contributions of this work are: 
\begin{enumerate} 
\item The introduction of \textbf{SAMSA 2.0}, which utilizes \emph{spectral angle prompting} to deliver higher accuracy with lower memory overhead. 
\item A comprehensive validation across an expanded collection of HSI datasets, demonstrating generalizability across different tissues and imaging systems. 
\item The adaptation of automated regional proposals for HSI medical images to accelerate clinical annotation. 
\end{enumerate}

\section{Related Work} \label{sec:related_work}

\textbf{Interactive Segmentation in Medical Imaging:} Interactive methods have long supported tasks in MRI, ultrasound, and microscopy. Early techniques like region growing and graph cuts \cite{boykovgraph} used sparse inputs to guide segmentation but often struggled with the complex, non-linear boundaries found in HSI data. Recent work by Wang et al. \cite{wang2024scribblebasedinteractivesegmentationmedical} has further demonstrated the growing interest in user-guided hyperspectral segmentation to improve clinical efficiency.

\textbf{Foundation Models for Segmentation:} Models like SAM \cite{kirillov2023segment} offer strong zero-shot performance on natural images. While extensions such as Grounding SAM \cite{ren2024grounded} and MedSAM-Adapter \cite{wu2023medical} enable domain adaptation for medical RGB or grayscale images, these models cannot process high-dimensional spectral data directly, necessitating custom integration strategies.

\textbf{Spectral-Spatial Fusion:} Spectral comparison methods, such as spectral angle \cite{boardman1993spectral} and Pearson correlation \cite{Meneses2000SCM}, are generalizable and require no training, but they lack spatial context and suffer from shading variability. Joint spectral-spatial models aim to combine these strengths. While the original SAMSA \cite{SAMSA} fused spectral maps with spatial features, it did so late in the architecture. SAMSA 2.0 represents a shift in this paradigm by embedding spectral prompts at the input level, enabling spatially-aware spectral feature extraction from the very first layer.

\section{Methodology}\label{sec:methodology}

\begin{figure}[t]
\centering
\includegraphics[width=0.9\linewidth]{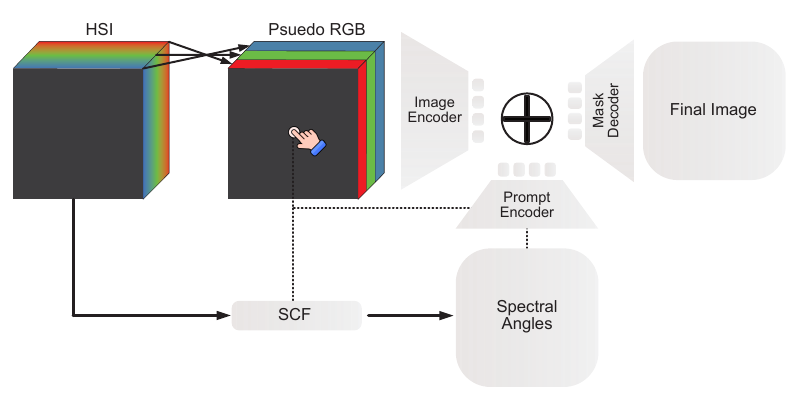}
\caption{SAMSA 2.0 Architecture featuring the early fusion of spectral similarity maps into the prompt encoder.}\label{fig:architecture}
\end{figure}

Interactive segmentation models operate by incorporating both input imagery and spatial cues provided by a user. In the Segment Anything Model (SAM) \cite{kirillov2023segment}, these cues typically include points, bounding boxes, or masks. While traditional spectral comparison methods rely solely on point-based input to identify spectrally similar regions, the original SAMSA framework \cite{SAMSA} introduced a hybrid approach by leveraging spectral similarity to guide spatial models. In this work, we build upon this concept by integrating spectral angle information significantly earlier in the pipeline. Rather than applying spectral analysis as a post-hoc refinement, we utilize the pretrained prompt encoder within SAM 2 to inject spectral similarity directly into the input representation. 

\subsection{Problem Formulation and Data Representation}
The objective is to delineate foreground regions from the background using minimal user input while exploiting high-dimensional spectral data. Let $X \in \mathbb{R}^{H \times W \times C}$ denote a hyperspectral image with spatial dimensions $H \times W$ and $C$ spectral bands. We derive a pseudo-RGB projection $X_{rgb} \in \mathbb{R}^{H \times W \times 3}$ from $X$. Ground truth segmentation is defined as $Y \in \{0, \dots, N\}^{H \times W}$, where $N$ is the number of semantic classes. User input is represented as a set of spatial cues $\mathcal{I} = \{I_{i,j}\}$, where each $I_{i,j} = (i, j)$ denotes a user-provided click at pixel location $(i, j)$. The model produces a binary probability mask $\hat{Y} \in [0, 1]^{H \times W}$.

\subsection{SAMSA 2.0: Early Spectral-Spatial Fusion}\label{sec:samsa2-arch}
The primary architectural innovation of SAMSA 2.0 is the integration of spectral information at the input of the prompt encoder. Given a hyperspectral image $X$ and click locations $\mathcal{I}$, we first compute a spectral similarity map $\hat{Y}_{\text{SCF}}$ using the Spectral Angle (SA) metric. To enhance regional separability and contrast, we apply histogram equalization to the resulting map, producing $\hat{Y}_{\text{SCF}}^{\text{eq}}$. This map is resized to match the resolution of $X_{rgb}$ and concatenated channel-wise. The resulting composite input is passed to the SAM 2 prompt encoder:
$$\hat{Y}_{\text{SAMSA 2.0}} = \text{SAM 2}(X_{rgb}, \hat{Y}_{\text{SCF}}^{\text{eq}}, \mathcal{I}).$$

\subsection{Training Protocol and Optimization}
We employ a per-class optimization scheme with multi-click gradient accumulation. For each semantic class $c$, we iteratively simulate user interaction:
$$\mathcal{I}^{(k)} = \{(i_0, j_0), (i_1, j_1), \dots, (i_{k-1}, j_{k-1})\},$$
where each new click is placed on the largest error region of the previous prediction. At each step $k$, the loss is computed as a weighted combination of binary cross-entropy ($\mathcal{L}_{\text{BCE}}$) and soft Dice loss ($\mathcal{L}_{\text{DSC}}$):
$$\mathcal{L}^{(k)} = \lambda \cdot \mathcal{L}_{\text{DSC}}^{(k)} + (1 - \lambda) \cdot \mathcal{L}_{\text{BCE}}^{(k)}.$$

\subsection{Automated Proposal Generation}
We adapt the automatic mask generation pipeline from SAM 2, prioritizing masks from smaller, higher-resolution crops during Non-Maximal Suppression (NMS). To address HSI challenges, we implement:
\begin{itemize}
\item \textbf{Lightweight Processing:} Fixed thresholding and NMS based on mask confidence.
\item \textbf{Heavy Processing:} Morphological closing, hole filling, and small connected component deletion.
\end{itemize}

\section{Results}

For optimization of trainable models we followed the same procedure as described in \cite{SAMSA}. At training time for the mixed training evaluation we employed a weighted random sampling strategy to address the inherent data imbalance across the constituent datasets. Using the sampler, data points were drawn with replacement based on pre-calculated sample weights. This oversampling technique ensures that each training batch contains a balanced representation from all four data sources (HIB, HEIPOR, SB-X, and SB-H), thereby preventing the model from becoming biased towards the larger datasets during optimisation.

\subsection{Datasets}

For validation we follow SAMSA's datasets HIB and HEIPOR, which are described in \cite{SAMSA}. In addition, we included the SB database~\cite{martin2025slimbrain}, consisting of two subsets: SB-X and SB-H. The SB-X subset was acquired with the Ximea snapshot camera, which captures 25 spectral bands in the 660--950~nm range. In contrast, the SB-H subset was recorded with the Headwall Hyperspec\textsuperscript{\textregistered} VNIR E-Series linescan camera, providing 369 effective spectral bands spanning 400--1000~nm. Together, these subsets offer complementary trade-offs between spectral coverage and resolution, enabling us to evaluate models under different imaging conditions. Both SB datasets contain five classes; Healthy, Blood, Tumour, Meninges and Unlabelled.

\subsection{Individual Dataset Study}

\begin{table}[ht]
\centering
\caption{
Performance comparison on HEIPOR, HIB, SB-X, and SB-H Datasets using SAM2 variants. 
D. denotes DICE; 1c/5c denote user clicks. 
Model suffixes: T (Tiny), S (Small), B+ (Base+), L (Large), FT (Fine-Tuned), Eq (Equalized), Ad (Adapter).
}\label{table:combined}
\footnotesize % More readable than the previous scaled-down version
\setlength{\tabcolsep}{2.5pt} % Tightened further to accommodate the '@' symbol

\begin{tabular}{lccc|ccc|ccc|ccc}
\toprule
\multirow{2.5}{*}{Model} 
& \multicolumn{3}{c|}{HEIPOR} 
& \multicolumn{3}{c|}{HIB} 
& \multicolumn{3}{c|}{SB-X} 
& \multicolumn{3}{c}{SB-H} \\
\cmidrule(lr){2-4} \cmidrule(lr){5-7} \cmidrule(lr){8-10} \cmidrule(lr){11-13}
& \shortstack{D.@.5\\(1c)} & \shortstack{D.@Mx\\(1c)} & \shortstack{D.@.5\\(5c)}
& \shortstack{D.@.5\\(1c)} & \shortstack{D.@Mx\\(1c)} & \shortstack{D.@.5\\(5c)}
& \shortstack{D.@.5\\(1c)} & \shortstack{D.@Mx\\(1c)} & \shortstack{D.@.5\\(5c)}
& \shortstack{D.@.5\\(1c)} & \shortstack{D.@Mx\\(1c)} & \shortstack{D.@.5\\(5c)} \\
\midrule
\multicolumn{13}{c}{\textbf{Spectral}} \\
\midrule
PCC & 12.2 & 47.2 & 11.7 & 37.3 & 88.5 & 37.5 & 31.4 & 52.7 & 29.1 & 37.4 & 77.5 & 37.4 \\
SA & 11.7 & 48.9 & 11.7 & 37.4 & 88.9 & 37.4 & 28.7 & 53.5 & 28.7 & 37.4 & 81.0 & 37.4 \\
SA$_{Eq}$ & 20.5 & 48.7 & 13.7 & 56.8 & 88.5 & 48.2 & 37.6 & 52.5 & 31.5 & 48.7 & 80.4 & 42.5 \\
\midrule
\multicolumn{13}{c}{\textbf{RGB}} \\
\midrule
SAM2-T & 64.2 & 79.3 & 64.8 & 54.4 & 74.8 & 59.6 & 47.9 & 68.5 & 56.1 & 67.0 & 84.6 & 65.4 \\
SAM2-T-FT & 79.6 & 85.7 & 88.8 & 81.2 & 91.3 & 92.2 & 69.5 & 88.0 & 88.7 & 82.5 & \textbf{92.9} & \textbf{93.7} \\
SAM2-S & 62.7 & 78.6 & 64.5 & 55.6 & 75.4 & 64.6 & 49.4 & 66.7 & 62.5 & 65.1 & 83.1 & 65.8 \\
SAM2-S-FT & 79.9 & 85.1 & 88.9 & 80.8 & 91.2 & 93.9 & 68.9 & 87.6 & 89.1 & 82.6 & 91.3 & 92.7 \\
SAM2-B+ & 61.7 & 76.6 & 60.1 & 53.8 & 71.2 & 60.9 & 43.5 & 60.3 & 56.7 & 57.5 & 76.4 & 59.5 \\
SAM2-B+-FT & 80.4 & 85.2 & 89.2 & 80.0 & 91.1 & 92.5 & 69.4 & 87.4 & 87.4 & 82.1 & 92.0 & 91.8 \\
SAM2-L & 60.0 & 77.3 & 64.4 & 52.3 & 72.7 & 59.1 & 53.0 & 66.7 & 61.1 & 62.6 & 80.9 & 64.8 \\
SAM2-L-FT & 80.2 & 86.0 & 89.2 & 74.9 & 90.3 & 91.3 & 70.7 & 85.4 & 90.0 & 83.4 & 90.6 & 91.5 \\
MedSAM-Ad. & 50.0 & 69.1 & 71.7 & 55.1 & 71.8 & 71.1 & 55.8 & 74.0 & 72.4 & 70.3 & 85.1 & 86.2 \\
\midrule
\multicolumn{13}{c}{\textbf{Fusion}} \\
\midrule
SAMSA & 80.4 & 85.9 & 88.8 & 81.8 & 93.2 & 93.5 & 71.0 & \textbf{89.9} & 90.1 & 82.7 & 91.3 & 92.9 \\
SAMSA2-T & 80.5 & 86.0 & 89.0 & 86.0 & \textbf{94.1} & 95.5 & \textbf{73.4} & 88.4 & 90.1 & \textbf{85.2} & 92.2 & 92.2 \\
SAMSA2-S & \textbf{82.2} & \textbf{86.5} & \textbf{89.7} & 86.2 & 93.6 & \textbf{96.4} & 72.2 & 89.5 & \textbf{90.6} & 81.8 & 90.6 & 91.0 \\
SAMSA2-B+ & 82.0 & 86.5 & 89.1 & \textbf{86.2} & 93.7 & 96.2 & 71.9 & 89.5 & 89.9 & 84.3 & 92.6 & 92.5 \\
SAMSA2-L & 81.4 & 86.2 & 88.7 & 85.0 & 93.8 & 94.6 & 72.4 & 88.8 & 90.2 & 84.2 & 91.4 & 92.2 \\
\bottomrule
\end{tabular}
\end{table}

Table \ref{table:combined} compares segmentation performance across four hyperspectral medical datasets, HEIPOR and HIB, SB-X, SB-H using spectral baselines, RGB-based models, and our proposed fusion methods. Spectral-only approaches (e.g., PCC, SA) show low DICE@0.5 scores due to the lack of spatial modeling but achieve high Max DICE scores, indicating the potential of spectral cues when thresholds are well-tuned. Fine-tuned RGB models (e.g., SAM2-Small-FT) perform better by leveraging spatial features, but still lack spectral awareness.
MedSAM-Adapter, in general, has pretty poor performance. We assume that the model has too many trainable weights for the size of the dataset.
SAMSA2 variants consistently outperform both spectral and RGB models. On average, SAMSA2-Small achieves a DICE@0.5 (1-click) score of 84.2\%, compared to 81.1\% for SAMSA and 80.4\% for the best RGB-only model. This corresponds to an average improvement of +3.1\% over SAMSA and +3.8\% over RGB-only segmentation.
In the 5-click setting, SAMSA2-Small achieves the highest DICE score of 96.4\% on the HIB dataset, demonstrating strong performance with minimal user input.
Note that SAMSA results differ from the original paper due to a modified training strategy. To reduce GPU memory, we accumulated gradients over user clicks only, rather than over both clicks and class labels.

\subsection{Mixed Training}

\begin{table*}[ht]
\centering
\caption{
Results across datasets with DICE at 0.5 threshold (D@0.5) at 1 (1c) and 5 (5c) clicks. 
Model suffixes: T (Tiny), S (Small), B (Base), L (Large), FT (Fine-Tuned). 
Values are percentages; bold indicates the best performance per column.
}\label{table:mixed_training}
\small % Reliable font size for 11 columns
\setlength{\tabcolsep}{4.5pt} % Balanced spacing for readability

\begin{tabular}{l cc|cc|cc|cc|cc}
\toprule
\multirow{2.5}{*}{Model} 
& \multicolumn{2}{c|}{HEIPOR} 
& \multicolumn{2}{c|}{HIB} 
& \multicolumn{2}{c|}{SB-X} 
& \multicolumn{2}{c|}{SB-H} 
& \multicolumn{2}{c}{Average} \\
\cmidrule(lr){2-3} \cmidrule(lr){4-5} \cmidrule(lr){6-7} \cmidrule(lr){8-9} \cmidrule(lr){10-11}
& \shortstack{D@.5\\(1c)} & \shortstack{D@.5\\(5c)} 
& \shortstack{D@.5\\(1c)} & \shortstack{D@.5\\(5c)} 
& \shortstack{D@.5\\(1c)} & \shortstack{D@.5\\(5c)} 
& \shortstack{D@.5\\(1c)} & \shortstack{D@.5\\(5c)} 
& \shortstack{D@.5\\(1c)} & \shortstack{D@.5\\(5c)} \\
\midrule
\multicolumn{11}{c}{\textbf{RGB}} \\
\midrule
SAM2-T-FT & 27.6 & \textbf{61.2} & 84.2 & 91.6 & \textbf{77.6} & 91.0 & 70.6 & 81.3 & 65.0 & 81.3 \\
SAM2-S-FT & 21.8 & 50.7 & 81.4 & 94.9 & 72.1 & 91.3 & 72.0 & 82.0 & 61.8 & 79.7 \\
SAM2-B-FT & 18.1 & 40.7 & 78.0 & 94.8 & 75.1 & 87.1 & 70.9 & 80.9 & 60.5 & 75.9 \\
SAM2-L-FT & \textbf{32.3} & 58.5 & 77.5 & 96.5 & 70.6 & 85.3 & 72.9 & 81.1 & 63.3 & 80.3 \\
\midrule
\multicolumn{11}{c}{\textbf{Fusion}} \\
\midrule
SAMSA2-T-FT & 23.4 & 54.5 & 87.0 & 96.8 & 73.7 & 91.8 & 71.8 & 82.8 & 64.0 & 81.5 \\
SAMSA2-S-FT & 20.0 & 57.3 & 86.6 & 95.6 & 72.9 & \textbf{93.4} & \textbf{73.5} & 82.5 & 63.3 & \textbf{82.2} \\
SAMSA2-B-FT & 23.4 & 48.0 & 83.9 & 96.4 & 73.1 & 90.4 & 73.1 & \textbf{83.2} & 63.4 & 79.5 \\
SAMSA2-L-FT & 30.7 & 53.9 & \textbf{89.9} & \textbf{97.8} & 74.9 & 89.8 & 72.7 & 80.9 & \textbf{67.1} & 80.6 \\
\bottomrule
\end{tabular}
\end{table*}

The effectiveness of our mixed-training strategy is demonstrated in \cref{table:mixed_training}. This table compares the post-fine-tuning performance of a standard RGB model (SAM2) against our hyperspectral-enabled fusion model (SAMSA2), both of which were subjected to the same mixed-training protocol.

The results indicate that, in general, mixed-training strategy provides a greater performance uplift to the hyperspectral-enabled SAMSA2 architecture. There is much variation in the "best performing model" per dataset, which we believe is due to different sized features for different datasets. This coupled with the sampling strategy could be a source for the variation. For the initial automated delineation from a single click, the SAMSA2-Large-FT model achieved the highest average DICE score of $67.1\%$. During the interactive refinement phase with five clicks, the SAMSA2-Small-FT model was most effective, reaching an average score of $82.2\%$. This suggests that our training approach is particularly successful at leveraging the richer data provided by the hyperspectral model for both initial contouring and subsequent adjustments.

However, performance gains in the 5-click (5c) scenario are less pronounced than in the 1-click case, with standard SAM2 remaining competitive, especially on HEIPOR. This likely reflects a limitation of our current multi-click mechanism, which updates prompts by taking the pixel-wise maximum across SA maps—an approach that may not optimally integrate multiple clicks and needs re-visiting.

On the HEIPOR dataset, the mixed training did not yield a performance advantage for the SAMSA2 model. This may indicate that for tissues with characteristics already distinct in the visible spectrum, our training method could not extract additional diagnostic value from the hyperspectral data. However, it is more likely that the sampling strategy is far too strong since the comparative size for HEIPOR to the other datasets is much larger.

In contrast, the benefits of applying our mixed-training protocol to the hyperspectral model were clearly evident on the HIB, SB-X, and SB-H datasets. The clinical utility was especially notable on the HIB dataset, where the trained SAMSA2-Large-FT model achieved a DICE score of $97.8\%$ after five clinician annotations, underscoring the success of our strategy in complex cases.

\begin{figure}[hp!]
    \centering
    \includegraphics[width=0.9\linewidth]{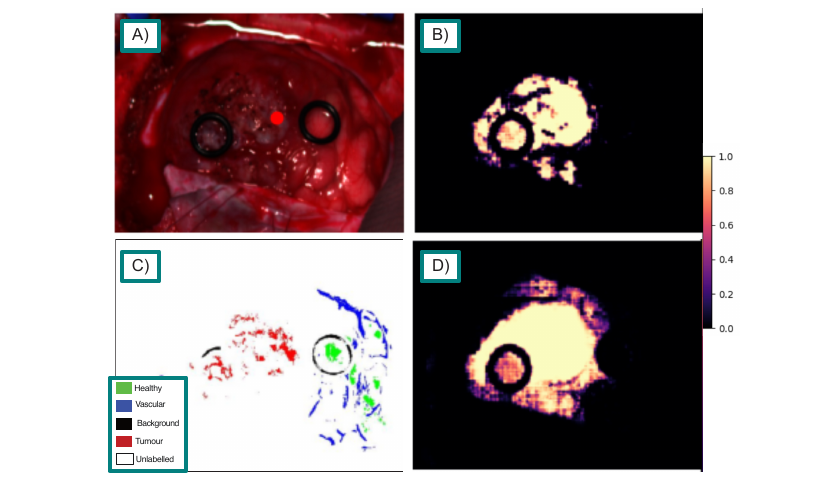}
    \caption{Sample segmentation results on the HIB dataset, tumour class. 
    A) Pseudo-RGB image of the target. 
    B) SAMSA2-Large standard predicted similarity.
    C) Ground truth labels. 
    D) SAM2-Large-FT using SAM2 predicted similarity}
    \label{fig:hib_tumour}
\end{figure}

\begin{figure}[hp!]
    \centering
    \includegraphics[width=0.9\linewidth]{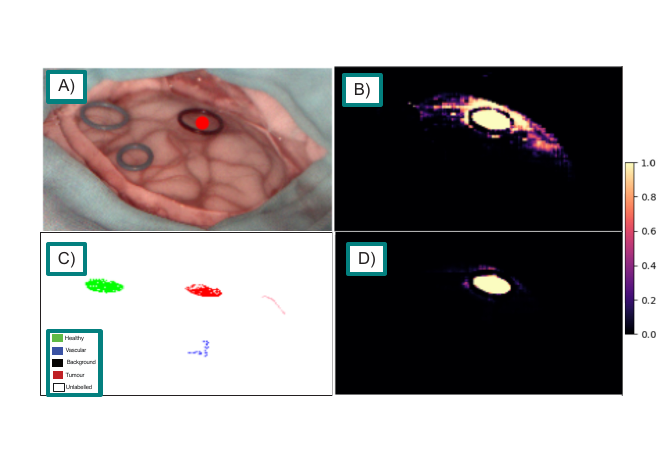}
    \caption{Sample segmentation results on the SB-X dataset, tumour class. 
    A) Pseudo-RGB image of the target. 
    B) SAMSA2-Large standard predicted similarity.
    C) Ground truth labels. 
    D) SAM2-Large-FT using SAM2 predicted similarity}
    \label{fig:sb-x-tumour}
\end{figure}

\begin{figure}[hp!]
    \centering
    \includegraphics[width=0.9\linewidth]{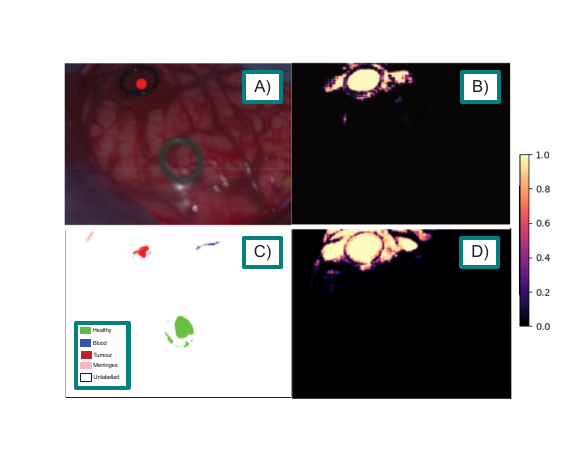}
    \caption{Sample segmentation results on the SB-H dataset, tumour class. 
    A) Pseudo-RGB image of the target.
    B) SAMSA2-Large standard predicted similarity.
    C) Ground truth labels. 
    D) SAM2-Large-FT using SAM2 predicted similarity}
    \label{fig:sb-h-tumour}
\end{figure}

\begin{figure}[hp!]
    \centering
    \includegraphics[width=0.9\linewidth]{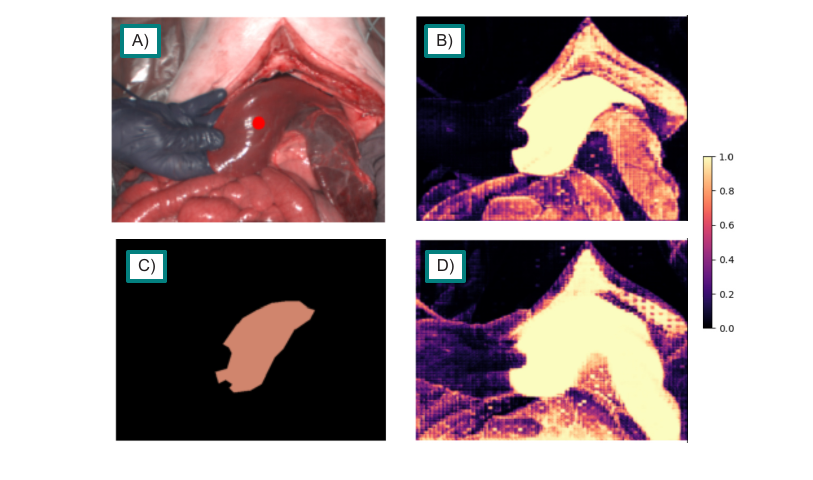}
    \caption{Sample segmentation results on the HEIPOR dataset, liver class. 
    A) Pseudo-RGB image of the target. 
    B) SAMSA2-Large standard predicted similarity.
    C) Ground truth labels. 
    D) SAM2-Large-FT using SAM2 predicted similarity}
    \label{fig:heipor_liver}
\end{figure}

\begin{figure}[htbp!]
    \centering
    \includegraphics[width=0.9\linewidth]{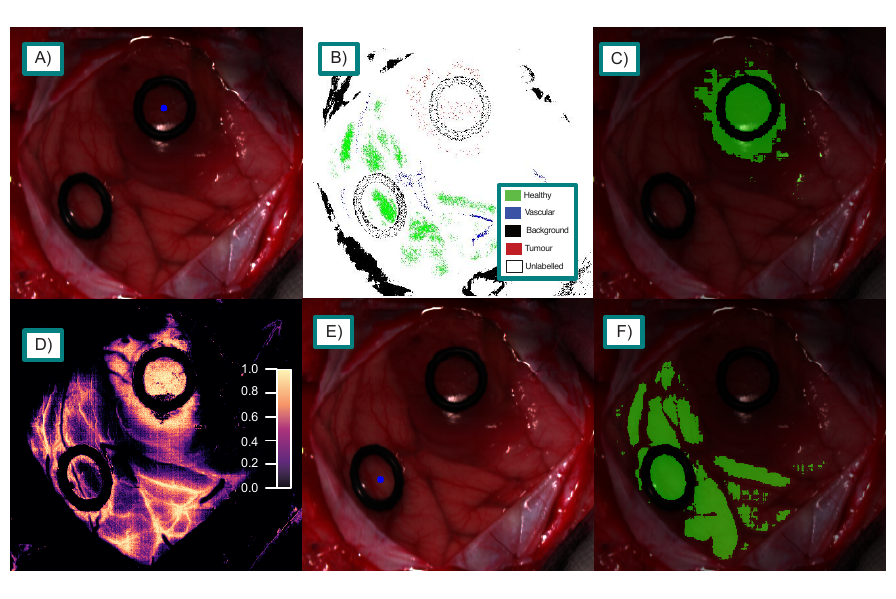}
    \caption{Example use case on HIB dataset, finding the boundary of tumour and non-tumour.
    A) Pseudo-RGB image of the target with tumour prompt. 
    B) Ground truth labels. 
    C) SAMSA2-Large standard segmentation from a).
    D) $SA_{Equalized}$ spectral similarity from a).
    E) Pseudo-RGB image of the target with healthy prompt.
    F) SAM2-Large-FT using SAM2 predicted similarity}
    \label{fig:hib_tumour_no_tumour}
\end{figure}

Qualitative comparisons further highlight the advantages of SAMSA2 over fine-tuned RGB baselines. On the HIB dataset (\cref{fig:hib_tumour}), both SAMSA2-Large and SAM2-Large-FT correctly localise the tumour, but SAMSA2 provides a more refined boundary, offering greater clinical interpretability. On the SB-X and SB-H datasets (\cref{fig:sb-x-tumour,fig:sb-h-tumour}), both methods capture the tumour within the marked region, yet SAMSA2 extends its prediction beyond the visible margins, suggesting the potential to reveal subsurface infiltration that may not be apparent in RGB-based predictions alone. For the HEIPOR dataset (\cref{fig:heipor_liver}), performance is broadly comparable, though SAMSA2 produces fewer false positives, avoiding spurious activations in neighbouring tissue seen with SAM2-Large-FT. Finally, \cref{fig:hib_tumour_no_tumour} illustrates a representative workflow with SAMSA2-Large: tumour prompts reveal infiltration extending beyond the visible tumour boundary, while healthy prompts delineate the transition from normal to abnormal tissue. This bidirectional prompting demonstrates how SAMSA2 could support decision-making by helping identify resection margins more precisely.

Finally, our mixed-training experiment reveals a nuance regarding model size and its role in the clinical workflow. The training benefited the larger SAMSA2-Large model most for the initial segmentation proposal. In contrast, the smaller SAMSA2-Small model was more responsive and effective when refining contours with user input, suggesting a potential to investigate model variants further.

\clearpage
\subsection{Automatic Segmentation}

\begin{figure}[hp!]
    \centering
    \includegraphics[width=\linewidth]{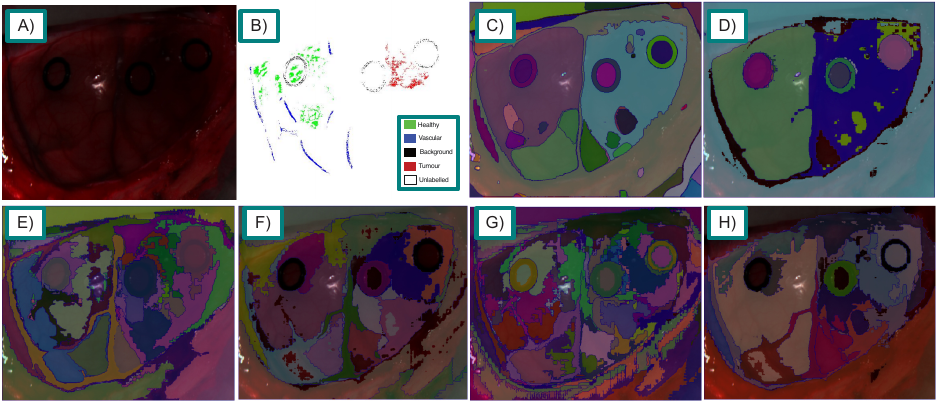}
    \caption{Automatic segmentation results on the HIB dataset. 
    A) Pseudo-RGB image of the target. 
    B) Ground truth labels. 
    C) SAM2-Large standard automatic segmentation. 
    D) SAM2-Large-FT using SAM2 standard automatic segmentation as input. 
    E) SAM2-Large-FT using our light automatic segmentation. 
    F) SAM2-Large-FT using our heavy automatic segmentation. 
    G) SAMSA2-Large-FT using our light automatic segmentation. 
    H) SAMSA2-Large-FT using our heavy automatic segmentation.}
    \label{fig:autoseg}
\end{figure}

\begin{figure}[hp!]
    \centering
    \includegraphics[width=\linewidth]{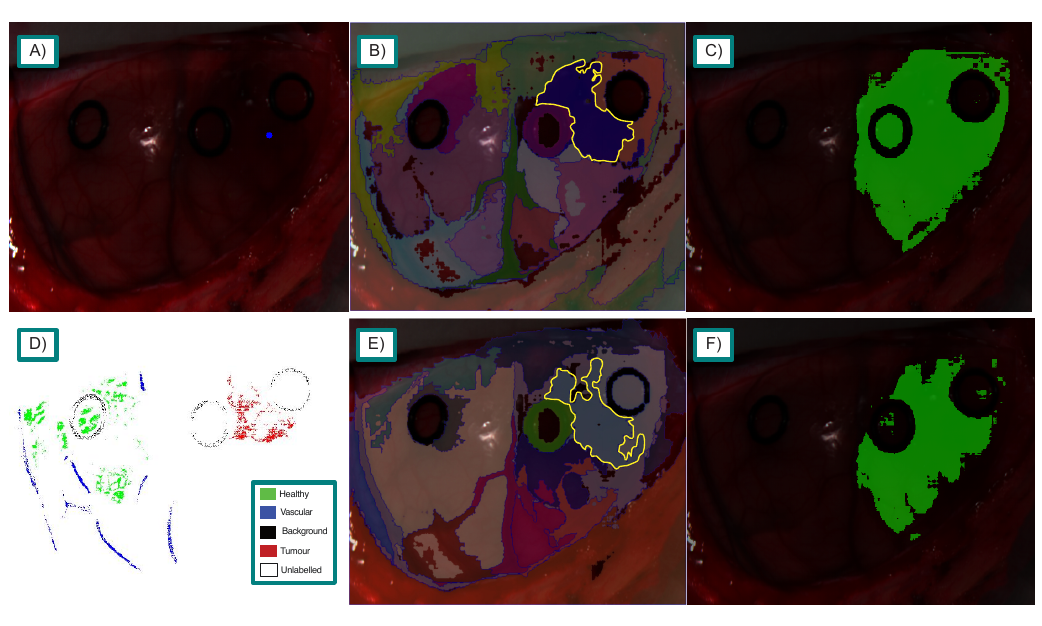}
    \caption{Example of interaction following automatic segmentation. 
    A) Pseudo-RGB image of the target, with a blue click point. 
    B) SAM2-Large-FT using our heavy automatic segmentation; the target-selected region is highlighted in yellow. 
    C) Output segmentation from SAM2-Large-FT using the prompt in (a). 
    D) Ground truth labels. 
    E) SAMSA2-Large-FT using our heavy automatic segmentation; the target-selected region is highlighted in yellow. 
    F) Output segmentation from SAMSA2-Large-FT using the prompt in (a).}
    \label{fig:autoseg2}
\end{figure}

Validating automatic segmentations generated by an interactive model remains an experimental task. The authors of SAM2 \cite{kirillov2023segment} rely on expert annotators and qualitative inspection for large-scale model training. We adopt a similar approach in this work.
In \cref{fig:autoseg}, we present the outputs of our automatic segmentation pipelines: SAM2 standard, as well as our light and heavy post-processing variations. The input RGB image is shown in \cref{fig:autoseg}a, with the corresponding ground truth label in \cref{fig:autoseg}b. The label reveals a clear hotspot or cluster of tumour annotations located between the two rings, representing the target region for our example surgical workflow.
SAM2-Large using the standard automatic segmentation pipeline is shown in \cref{fig:autoseg}c. While the segmentation is smooth and effectively identifies macro-structures, it lacks clinical relevance. The model captures large brain fold regions, which are disconnected by vessels, and distinguishes between the interior and material of the rings. To demonstrate that these limitations stem from the automatic segmentation pipeline rather than the model itself, \cref{fig:autoseg}d shows SAMSA2-Large-FT with the standard pipeline. The segmentation maps appear noisier, as expected following fine-tuning on medical tissue, and overall performance is similar or slightly worse than SAM2-Large.
Applying our light and heavy post-processing pipelines to SAM2-Large-FT results in the segmentations shown in \cref{fig:autoseg}e and \cref{fig:autoseg}f. The light variation produces noisier outputs but still highlights patchy regions that the model identifies as distinct. The heavy variation yields smoother, more contiguous masks.
These results suggest that simplifying SAM2's complex automatic segmentation pipeline can produce more clinically meaningful masks. However, the outputs remain noisy, and the shapes of segmented regions are often irregular, which may limit direct clinical use without further refinement.
To illustrate potential applications, \cref{fig:autoseg2} presents an example workflow. Suppose a surgeon identifies a patchy region surrounding the tumour as clinically relevant. This region can be selected interactively, as shown in \cref{fig:autoseg2}b and e. A single centre-of-mass click can then be used to generate targeted segmentations, as illustrated in \cref{fig:autoseg2}. The resulting segmentations from SAM2-Large-FT and SAMSA2-Large are shown in \cref{fig:autoseg2}c and f, respectively.
This workflow has potential to support surgical decision-making by allowing rapid, interactive identification of regions of interest. For example, a surgeon could quickly highlight suspicious tissue during preoperative planning, receive an automatically generated segmentation of the surrounding anatomy, and adjust surgical strategy accordingly. Such a system could also be used intraoperatively to guide resection margins, identify critical structures, or provide real-time visual feedback, ultimately improving both safety and precision in neurosurgical procedures.

\section{Clinical Usage of SAMSA2.0}
The clinical usage follows the main themes described in \cite{SAMSA} but we have no enabled automatic mask generation. Automatic mask generation shifts segmentation from fully interactive to hybrid autonomous operation. This reduces reliance on intraoperative input while still allowing surgeon guidance when needed. Automatic segmentation can run continuously, identifying tissue types without disrupting workflow.
The dual-mode design supports different surgical needs: surgeons may use interactive mode for precise tumor margin delineation, while routine monitoring can remain automated.
Although automation may introduce noise or irregular boundaries, lightweight post-processing produces clinically usable segmentations. This balance between speed and precision reflects surgical practice, where approximate boundaries often suffice for decision-making.
The key question is how surgeons adopt the system in practice. A focused clinical study should assess how surgeons integrate SAMSA2 into their workflow. Surgeons would perform standardized tasks using both automatic and interactive modes, with data collected on decision time, override frequency, and segmentation accuracy. Usability, workload, and trust surveys, along with post-session interviews, would capture acceptance and preferred mode balance. Such a study requires fully labelled datasets rather than sparsely annotated ones like the HIB dataset to ensure reliable evaluation.

An additional consideration for eventual clinical and commercial deployment is the licensing of the underlying foundation models. The SAM2 model weights are released under the Apache 2.0 license, which permits modification, redistribution, and commercial use with minimal restrictions. This is advantageous for medical applications, where regulatory approval and commercial translation often require the ability to integrate, adapt, and redistribute models within proprietary systems. In contrast, some alternative foundation models impose more restrictive licenses that limit commercial usage or derivative works, potentially hindering clinical adoption. The permissive licensing of SAM2 therefore supports the long-term feasibility of deploying SAMSA in real-world clinical and commercial settings.

\section{Conclusion}

We presented SAMSA2, a method that enhances hyperspectral segmentation by embedding spectral angle information as a spatially aware input prompt. This early fusion strategy improves robustness in low-data and noisy conditions, consistently outperforming both spectral-only and RGB-only baselines with minimal user input. In addition, we explored automatic mask generation pipelines, demonstrating that lightweight post-processing can yield clinically meaningful segmentations, though at times with increased noise or irregular boundaries.
Looking forward, further gains may come from end-to-end learning with the mask decoder, expanding training on larger and more diverse datasets, and validating performance in prospective clinical workflow studies. Ultimately, integrating SAMSA2 into intraoperative environments could provide surgeons with real-time, interactive guidance, supporting precise identification of tumour margins and improving decision-making during procedures.

\clearpage
\bibliographystyle{plain}
\bibliography{references}

\end{document}